\documentclass[11pt]{article}
\usepackage{amsmath,amssymb,amsthm}
\usepackage[english]{babel}
\usepackage{graphicx}
\usepackage{xcolor}
\usepackage[authoryear]{natbib}
\usepackage{authblk}
\usepackage{etoolbox}
\usepackage{flafter}
\usepackage{needspace}
\usepackage{placeins}
\usepackage[ruled,vlined,algo2e]{algorithm2e}
\usepackage[breaklinks=true]{hyperref}

\allowdisplaybreaks

\bibpunct[, ]{(}{)}{,}{a}{}{,}

\hypersetup{
  colorlinks=true,
  linkcolor={red!50!black},
  citecolor={blue!50!black},
  urlcolor={blue!80!black},
  pdftitle={Optimal Online Regret for Context-Lumpable Bandits},
  pdfauthor={Chenyu Lu and Zihan Zhang}
}

\newtheorem{theorem}{Theorem}
\newtheorem{lemma}{Lemma}
\newtheorem{corollary}{Corollary}

\SetAlgoCaptionSeparator{\ }
\AtBeginDocument{\SetAlgoVlined\DontPrintSemicolon}
\SetAlFnt{\small}
\SetAlgoSkip{smallskip}
\AfterEndEnvironment{algocf@Here}{\par\addvspace{\textfloatsep}}

\AtBeginEnvironment{theorem}{\postdisplaypenalty=10000}
\AtBeginEnvironment{proof}{\postdisplaypenalty=10000}

\title{Optimal Online Regret for Context-Lumpable Bandits}
\author[1]{Chenyu Lu\thanks{Email: \texttt{cluba@connect.ust.hk}}}
\author[1]{Zihan Zhang}
\affil[1]{The Hong Kong University of Science and Technology}
\date{}

\begin{document}
\raggedbottom
\maketitle

\begin{abstract}
We study stochastic contextual bandits with an unknown context-lumpable structure. The $S$ contexts are partitioned into $r$ blocks, and every arm has the same mean reward across contexts in the same block. Under arbitrary context distributions and without separation assumptions, we design an algorithm with expected regret $
\widetilde O\left(\sqrt{r(S+K)T}\right)$,
matching the lower bound of \citet{lee2023context} up to logarithmic factors and improving their previous upper bound by a factor of $r$.

The algorithm operates over dyadic accuracy levels, using a baseline policy and confidence-based early stopping to control exploration. It then performs cross-context elimination: since rewards are identical within a block, one selected arm eliminates all surviving records from that block, requiring at most $r$ selections at each level. Multi-scale exploration assigns finer accuracies to frequent blocks and coarser accuracies to rare blocks, yielding a near-optimal arm for every block while keeping the weighted policy error small.
\end{abstract}

\section{Introduction}
\label{sec:introduction}

In contextual bandits, the learner observes a context, selects an arm, and receives feedback only for that arm. In recommendation, for example, feedback is observed only for displayed items \citep{li2010contextual}. Sharing observations across contexts can reduce learning cost, but the relation between their rewards may be unknown.

In context-lumpable stochastic bandits \citep{lee2023context}, $S$ contexts form $r\le\min\{S,K\}$ unknown blocks. Each of the $K$ arms has the same mean reward within a block. Observations can thus be shared within blocks, but the learner sees context identities without their block assignments. It must find useful arms and learn where they perform well from bandit feedback.
Running a separate bandit for each context gives expected regret $\widetilde O(\sqrt{SKT})$ over $T$ rounds \citep{lattimore2020bandit}, which can be wasteful when $r$ is small. Yet exploration across contexts incurs regret: an informative arm may be poor where it is sampled. Finding a good policy and minimizing regret are therefore distinct challenges.
\citet{lee2023context} established nearly optimal sample complexity for finding a near-optimal policy of $\widetilde{O}\left(\frac{r(S+K)}{\epsilon^2}\right)$, while a sub-optimal regret bound of $\widetilde O(\sqrt{r^3(S+K)T})$. Their regret algorithm refines a context partition while eliminating arms; ensuring that refinements do not split a true block incurs an extra factor of $r$. They left $\widetilde O(\sqrt{r(S+K)T})$ regret open. We attain this rate by controlling exploration during candidate-arm selection, without recovering the partition.

\subsection{Our Results}

For every context distribution $\nu$ and its induced block distribution $\omega$, our algorithm achieves expected regret
\[
    \operatorname{Reg}_T
    =
    \widetilde O\!\left(\sqrt{r(S+K)T}\right),
\]
where $\widetilde O(\cdot)$ hides polylogarithmic factors in $S$, $K$, $r$, and $T$. For a known horizon, this guarantee is achieved by Algorithm~4; a standard doubling restart gives the same order without knowing $T$.

This rate matches the $\Omega(\sqrt{r(S+K)T})$ lower bound of Lee et al.~(2023) up to logarithmic factors. It improves their online upper bound $\widetilde O(\sqrt{r^3(S+K)T})$ to $\widetilde O(\sqrt{r(S+K)T})$, removing an additional factor of $r$. The guarantee holds for arbitrary block and context distributions and does not require any separation between the reward vectors of different blocks. Conceptually, the algorithm exploits the equality of mean rewards within each block without first recovering the unknown context partition: it constructs a shared set of candidate arms that contains a sufficiently good arm for every block and then selects among these candidates separately at each context.

The first technical ingredient is a phased exploration procedure with geometrically decreasing accuracy levels. In phase $h$, the algorithm targets accuracy $\varepsilon_h$ and first estimates the preceding policy $\pi_{h-1}$ at every context, obtaining a context-wise baseline $B_h(i)$. It then runs the \textsc{Collect} procedure at levels $\varepsilon_n=2^{-n/2}$ for $n=0,\ldots,h$. At level $n$, each context receives $q_{h,n}\asymp  rK\varepsilon_n/(S\varepsilon_h)$ random arm assignments, and each assignment is sampled up to $m_{h,n}\asymp 1/\varepsilon_n^2$ times. Sampling stops as soon as the arm's upper confidence bound falls below the baseline by a margin of order $\varepsilon_n$. On the joint concentration event, every arm stopped in this way is worse than the preceding policy by a constant multiple of $\varepsilon_n$, whereas every arm that completes its budget has an $\varepsilon_n$-accurate estimate. Thus, the algorithm avoids repeatedly paying for assignments that have already been certified to be harmful, while the multiple accuracy levels allow it to search for good arms at a scale adapted to the frequency of each block.

The second ingredient is a multi-scale, cross-context elimination rule. At each level, the algorithm selects the surviving context-arm record with the largest estimated reward, evaluates the selected arm across all contexts, and removes every record for which the selected arm or the preceding policy is sufficiently better. Since all contexts in the selected arm's true block have the same mean reward, one selection removes all remaining records from that block; consequently, at most $r$ selections are needed at any level, even though the block partition is never identified. In the uniform-context analysis, define
\[
    \varepsilon_{h,b}
    =
    \max\!\left\{1,\frac{1}{r\omega(b)}\right\}\varepsilon_h .
\]
For every block with $\varepsilon_{h,b}\le 1$, the assignment budget ensures that its optimal arm is represented at some level satisfying $\varepsilon_{h,b}\le \varepsilon_n<\sqrt{2}\,\varepsilon_{h,b}$. The elimination step then leaves the candidate set with an arm whose gap on block $b$ is $O(\varepsilon_{h,b})$. The final finest-scale update, which retains $\pi_{h-1}$ as a fallback, therefore satisfies
\[
    \max_{j\in[K]} A(i,j)-A(i,\pi_h(i))
    \le
    \min\{1,10\varepsilon_{h,b}\},
    \qquad i\in B(b).
\]
Although small blocks are learned only at coarser accuracy, their errors are weighted by their smaller arrival probabilities:
\[
    \sum_{b=1}^r \omega(b)\varepsilon_{h,b}
    =
    \varepsilon_h\sum_{b=1}^r
    \max\!\left\{\omega(b),\frac1r\right\}
    \le 2\varepsilon_h .
\]
Hence the phase policy has average gap $O(\varepsilon_h)$, while early stopping and the at most $r$ eliminations per level keep the exploration cost at $\widetilde O(r(S+K)/\varepsilon_h)$. Balancing this cost with phases of length $\widetilde O(r(S+K)/\varepsilon_h^2)$ and summing over the geometric sequence of phases yields the stated regret bound. For an arbitrary context distribution, Algorithm~4 first groups contexts with comparable estimated arrival probabilities, applies the same procedure separately within each approximately uniform group, discards only a low-frequency set of small total probability mass, and combines the group guarantees by Cauchy--Schwarz inequality.
\subsection{Related Work}

\paragraph{Contextual and latent bandits.}
Linear contextual bandits share reward parameters through known features \citep{chu2011contextual}, whereas \citet{maillard2014latent} studied latent bandits with clustered rewards. Online clustering for linear bandits has been studied under parameter separation \citep{gentile2014online} and with item-dependent groups \citep{gentile2017context}. \citet{kausik2025offline} learned a shared reward-parameter subspace from offline data for online linear latent contextual bandits. We assume neither such features nor offline data.

\paragraph{Bandits with identical reward vectors.}
\citet{pal2023optimal} combined clustering and matrix completion for identical within-cluster reward vectors under a uniform context distribution. Their near-optimal regret guarantee assumes incoherence, bounded condition number, and a column-subset condition; their main presentation treats the number of clusters and their size ratio as constants. \citet{lee2023context} studied context lumpability without these matrix conditions. We follow this setting, with explicit dependence on the number of blocks and arbitrary block and context distributions.

\paragraph{Low-rank bandits.}
A reward matrix with at most $r$ distinct rows has rank at most $r$, so context lumpability is a special case of low-rank structure. \citet{jun2019bilinear} studied bilinear bandits with known features. \citet{lu2021lowrank} and \citet{kang2022efficient} considered low-rank generalized linear bandits, where actions have known matrix features and the unknown parameter is low-rank. For linear rewards, \citet{jang2024efficient} used experimental design to obtain regret bounds that depend on the geometry of the arm set. Other work addresses heavy-tailed rewards \citep{kang2024heavy} and minimax lower bounds for low-rank matrix parameters \citep{zeng2025succinct}.

For low-rank reward matrices, \citet{stojanovic2023spectral} combined spectral estimation with arm elimination when the learner can choose any matrix entry. Closer to our setting, \citet{jedra2024lowrank} studied randomly arriving contexts using singular-subspace estimation and a reduction to misspecified linear bandits. Our analysis instead exploits equality of mean rewards within each block, a stronger assumption than low rank alone, without estimating a factorization or singular subspaces.

\section{Notation and problem formulation}\label{sec:setting}
Write $[n]=\{1,\ldots,n\}$, and take all logarithms to be natural. A context-lumpable bandit \citep{lee2023context} has $S$ contexts, $K$ arms, and $r$ nonempty blocks, where $1\le r\le\min\{S,K\}$. An unknown map $g:[S]\to[r]$ defines the blocks $\mathcal B(b)=\{i\in[S]:g(i)=b\}$. For an unknown $\mu:[r]\times[K]\to\mathbb R$, the mean reward matrix satisfies
\begin{equation*}
A(i,j)=\mu(g(i),j),\qquad (i,j)\in[S]\times[K].
\end{equation*}
The learner knows $S,K,r$, but not $g$ or $A$. No positive separation between the mean-reward vectors of distinct blocks is assumed.

At round $t$, a context $i_t$ is drawn from an unknown, fixed context distribution $\nu$ on $[S]$, independently of the preceding history. The learner observes $i_t$, chooses $j_t\in[K]$, and receives $y_t=A(i_t,j_t)+\eta_t$. We assume, almost surely,
\begin{equation}\label{eq:noise}
\begin{aligned}
\mathbb E[\eta_t\mid i_1,j_1,y_1,\ldots,i_t,j_t]&=0,\\
\mathbb E[e^{\lambda\eta_t}\mid i_1,j_1,y_1,\ldots,i_t,j_t]
&\le e^{\lambda^2/2}\quad(\lambda\in\mathbb R).
\end{aligned}
\end{equation}
Here and below, conditioning also includes the learner's randomness revealed before the current reward. Thus, $\eta_t$ is conditionally $1$-sub-Gaussian given the information available immediately before $y_t$.
The mean rewards have unit span:
\begin{equation}\label{eq:span}
\max_{j\in[K]}A(i,j)-\min_{j\in[K]}A(i,j)\le1,\qquad i\in[S].
\end{equation}
Consequently, every play incurs at most one unit of regret, including rounds used only for initial exploration.

The block distribution is given by $\omega(b)=\sum_{i\in\mathcal B(b)}\nu(i)$; it is unknown to the learner. Under a uniform context distribution, $\omega(b)=|\mathcal B(b)|/S$. We impose no restriction on the block distribution.

The arrival probability $\nu(i)$ controls how often an arm can be tested at context $i$, whereas $\omega(b)$ controls how often a choice on block $b$ affects expected regret. Under a uniform context distribution, a smaller block offers fewer opportunities to find a good arm, but choosing a suboptimal arm there also has less influence on regret. This balance motivates the use of multiple sampling accuracies below.

\Needspace{11\baselineskip}
A policy is a map $\pi:[S]\to[K]$. For each block $b$, fix $j_b^\star\in\arg\max_{j\in[K]}\mu(b,j)$. The expected regret over $T\ge1$ rounds is
\begin{equation*}
\operatorname{Reg}_T=\mathbb E\!\left[\sum_{t=1}^T\bigl(\max_{j\in[K]}A(i_t,j)-A(i_t,j_t)\bigr)\right].
\end{equation*}
The expectation includes contexts, rewards, and the learner's randomness. The notation $\widetilde O(\cdot)$ hides universal constants and factors polylogarithmic in $S,K,r,T$. Algorithm~\ref{alg:arbitrary-blocks} does not require $T$; Algorithm~\ref{alg:general} uses $T$, while Theorem~\ref{thm:general} also gives a horizon-free version.

\section{Regret under a Uniform Context Distribution}\label{sec:uniform-contexts}

\subsection{Online Algorithm}\label{sec:arbitrary-blocks}
We now describe our algorithm in detail. As shown in Algorithm~\ref{alg:arbitrary-blocks}, it proceeds in phases indexed by dyadic precision levels. Each phase consists of the following four steps.

\paragraph{Parameter settings.} Set $\varepsilon_n=2^{-n/2}$ for $n\ge0$. In phase $h\ge1$, we use $\beta_h=20\log(16SKr2^h)$, assignment counts $q_{h,n}=\lceil\beta_hrK\varepsilon_n/(S\varepsilon_h)\rceil$ and sample budgets $m_{h,n}=\lceil8\beta_h/\varepsilon_n^2\rceil$ for $n\in\{0,\ldots,h\}$, and $L_h(k)=\lceil128S(k+\beta_h)\rceil$. Write $m_h=m_{h,h}$ for the finest-level sample budget.

\textbf{Step 1.} We first estimate the mean reward of the preceding policy at each context using Baseline. These estimates provide a reference for Collect to stop sampling an arm when its reward is sufficiently below the baseline, avoiding unnecessary regret (Lemma~\ref{lem:b-baseline}).

\textbf{Step 2.} Since the block distribution is unknown, we explore all accuracy levels $n=0,\ldots,h$. Coarser levels assign more arms to each context but collect fewer rewards per arm, increasing the chance of sampling an optimal arm from a small block. Finer levels provide more accurate estimates. For each level, $D_{h,n}$ records the context-arm pairs that complete sampling without being stopped. This allows optimal arms from different blocks to be sampled at suitable levels (Lemmas~\ref{lem:b-means}--\ref{lem:b-draws}).

\begin{algorithm2e}[!htbp]
\caption{Online Algorithm for Arbitrary Block Distributions}\label{alg:arbitrary-blocks}
Let $t$ denote the current time and initialize $W_0\leftarrow\{1\}$, $\pi_0(i)\leftarrow1$ for $i\in[S]$\;
\For{phase $h=1,2,\ldots$}{
$t_h\leftarrow t$, $W_h\leftarrow W_{h-1}$\;
\textbf{Step 1. Baseline estimation}\;
$B_h\leftarrow\operatorname{Baseline}(h,\pi_{h-1})$\;
\textbf{Step 2. Random arm exploration}\;
\For{$n=0,\ldots,h$}{
$Q_{h,n}(i)\overset{\mathrm{i.i.d.}}{\sim}\operatorname{Unif}([K]^{q_{h,n}})$ for $i\in[S]$\;
$(D_{h,n},\widehat A_{h,n})\leftarrow\operatorname{Collect}(h,n,Q_{h,n},\pi_{h-1},B_h)$\;
}
\textbf{Step 3. Candidate selection}\;
\For{$n=0,\ldots,h$}{
\For{$\ell=1,\ldots,r$}{
\If{$D_{h,n}=\varnothing$}{\textbf{break}\;}
$(i^\star,j^\star)\in\arg\max_{(i,j)\in D_{h,n}}\widehat A_{h,n}(i,j)$,
$W_h\leftarrow W_h\cup\{j^\star\}$\;
Set $Q(i)=\{j^\star\}$ for all $i$;
$(D,\widehat A)\leftarrow\operatorname{Collect}(h,n,Q,\pi_{h-1},B_h)$\;
$\displaystyle\widehat V(i,j^\star)=\begin{cases}
\max\{B_h(i),\widehat A(i,j^\star)\},&(i,j^\star)\in D,\\
B_h(i),&(i,j^\star)\notin D,
\end{cases}\quad i\in[S]$\;
$D_{h,n}\leftarrow\{(i,j)\in D_{h,n}:\widehat V(i,j^\star)<\widehat A_{h,n}(i^\star,j^\star)-4\varepsilon_n\}$\;
}
}
\textbf{Step 4. Policy update}\;
Set $Q(i)=W_h$ for all $i$;
$(D,\widehat A)\leftarrow\operatorname{Collect}(h,h,Q,\pi_{h-1},B_h)$\;
$D\leftarrow D\cup\{(i,\pi_{h-1}(i)):i\in[S]\}$,
$\widehat A(i,\pi_{h-1}(i))\leftarrow B_h(i)$ for all $i$\;
$\pi_h(i)\leftarrow\arg\max_{j:(i,j)\in D}\widehat A(i,j)$ for all $i$\;
\While{$t-t_h<\left\lceil2^{16}\beta_h^3r(S+K)/\varepsilon_h^2\right\rceil$}{
Receive $i_t$; play $\pi_h(i_t)$\;
}
}
\end{algorithm2e}

\textbf{Step 3.} For each level, we select the arm $j^\star$ with the largest estimated reward in $D_{h,n}$ and evaluate it across all contexts. We remove records for which either $j^\star$ or the preceding policy achieves a reward sufficiently close to the selected estimate. Since contexts in the same block have identical mean rewards, each iteration removes all remaining records from at least one true block. Thus, at most $r$ iterations are needed per level, and the resulting candidate set $W_h$ contains a near-optimal arm for every block at an accuracy depending on its size, without recovering the block partition (Lemma~\ref{lem:b-candidates}).

\textbf{Step 4.} Finally, we evaluate the arms in $W_h$ at the finest level and select the arm with the largest estimated reward for each context, retaining the preceding policy as an option. Although smaller blocks may have larger estimation errors, they also appear less frequently. Consequently, the resulting policy satisfies
\[
\frac1S\sum_{i=1}^S\left[\max_j A(i,j)-A(i,\pi_h(i))\right]\le20\varepsilon_h
\]
on the event guaranteed by Lemma~\ref{lem:b-event} (Lemma~\ref{lem:b-policy}). Together with early stopping, this allows us to control the exploration regret in each phase and obtain the bound in Theorem~\ref{thm:blocks}.

\Needspace{6\baselineskip}
\subsection{Subroutines}\label{sec:subroutines}

The online algorithm uses two subroutines, Baseline and Collect, described below.

\textbf{Baseline.} In phase $h$, Baseline plays the preceding policy $\pi_{h-1}$ for $L_h(m_h)$ rounds. Under a uniform context distribution, this duration ensures that every context appears at least $m_h$ times with probability at least $1-Se^{-8\beta_h}$. Averaging the first $m_h$ observed rewards at each context $i$ gives $B_h(i)$, an estimate of $A(i,\pi_{h-1}(i))$. Since $m_h=\lceil8\beta_h/\varepsilon_h^2\rceil$, concentration guarantees
\[
|B_h(i)-A(i,\pi_{h-1}(i))|\le\varepsilon_h/2,\qquad i\in[S],
\]
with high probability (Lemma~\ref{lem:b-baseline}).

\begin{algorithm2e}[!htbp]
\caption{\textnormal{Baseline}$(h,\pi)$}\label{alg:baseline}
$n_i\leftarrow0$, $B_h(i)\leftarrow0$ for $i\in[S]$\;
\For{$\tau=1,\ldots,L_h(m_h)$}{
Receive $i_t$; play $\pi(i_t)$ and observe $y_t$\;
\lIf{$n_{i_t}<m_h$}{$B_h(i_t)\leftarrow\dfrac{n_{i_t}B_h(i_t)+y_t}{n_{i_t}+1}$, $n_{i_t}\leftarrow n_{i_t}+1$}
}
\Return{$B_h=(B_h(i))_{i\in[S]}$}\;
\end{algorithm2e}

\begin{algorithm2e}[!htbp]
\caption{\textnormal{Collect}$(h,n,Q,\pi,B_h)$}\label{alg:collect}
$D\leftarrow\varnothing$, $a_i\leftarrow1$, $n_i\leftarrow0$ for $i\in[S]$\;
$\widehat A_0(i,j)\leftarrow0$ for $(i,j)\in[S]\times[K]$\;
\For{$\tau=1,\ldots,L_h(m_{h,n}\max_i|Q(i)|)$}{
Receive $i_t$\;
\eIf{$a_{i_t}>|Q(i_t)|$}{
Play $\pi(i_t)$\;
}{
$j\leftarrow Q(i_t)_{a_{i_t}}$; play $j$ and observe $y_t$\;
$\widehat A_{n_{i_t}+1}(i_t,j)\leftarrow\dfrac{n_{i_t}\widehat A_{n_{i_t}}(i_t,j)+y_t}{n_{i_t}+1}$,
$n_{i_t}\leftarrow n_{i_t}+1$\;
\eIf{$j\ne\pi(i_t)$ and $\widehat A_{n_{i_t}}(i_t,j)+\sqrt{2\beta_h/n_{i_t}}<B_h(i_t)-3\varepsilon_n$}{
$a_{i_t}\leftarrow a_{i_t}+1$, $n_{i_t}\leftarrow0$\;
}{
\If{$n_{i_t}=m_{h,n}$}{
$D\leftarrow D\cup\{(i_t,j)\}$, $\widehat A(i_t,j)\leftarrow\widehat A_{m_{h,n}}(i_t,j)$\;
$a_{i_t}\leftarrow a_{i_t}+1$, $n_{i_t}\leftarrow0$\;
}
}
}
}
\Return{$(D,\widehat A)$}\;
\end{algorithm2e}

\textbf{Collect.} At level $n$, Collect samples the arms in $Q(i)$ one by one, with at most $m_{h,n}$ observations per arm. The stopping rule ensures that whenever an arm $j$ is stopped early at context $i$, $A(i,j)\le A(i,\pi_{h-1}(i))-\frac52\varepsilon_n$ on the event $\Omega_h$ (Lemmas~\ref{lem:b-baseline}--\ref{lem:b-means}). Thus, early stopping avoids further regret from arms that perform substantially worse than the preceding policy. Arms that are not stopped are recorded in $D$, together with their estimated mean rewards $\widehat A(i,j)$, after $m_{h,n}$ observations. The duration $L_h(m_{h,n}\max_i|Q(i)|)$ ensures that every context appears sufficiently often to process all arms in $Q(i)$, with probability at least $1-Se^{-8\beta_h}$ (Lemma~\ref{lem:b-completion}). Once all arms in $Q(i)$ have been processed, Collect plays $\pi_{h-1}(i)$ whenever context $i$ appears.

\subsection{Regret Analysis}
We now state our main result.

\begin{theorem}\label{thm:blocks}
Under a uniform context distribution $\nu$ and an arbitrary induced block distribution $\omega$, Algorithm~\ref{alg:arbitrary-blocks} requires no knowledge of $T$ and satisfies \mbox{$\operatorname{Reg}_T=\widetilde O\!\left(\sqrt{r(S+K)T}\right)$}.
\end{theorem}

\textbf{Proof sketch.} The key is to bound the regret incurred by Collect. If an arm is much worse than the preceding policy, Collect stops sampling it early. Otherwise, its reward is not much lower than that of the preceding policy. Since $\pi_{h-1}$ has an average gap of $O(\varepsilon_h)$ (Lemma~\ref{lem:b-policy}), sampling one arm at every context at level $n$ incurs at most $O(S\beta_h/\varepsilon_n)$ regret. Summing over the arms sampled in Steps 2--4 gives $O(\beta_h^3r(S+K)/\varepsilon_h)$.

The remaining rounds play either $\pi_{h-1}$ or $\pi_h$, both with average gap $O(\varepsilon_h)$ (Lemma~\ref{lem:b-policy}). Since phase $h$ lasts $O(\beta_h^3r(S+K)/\varepsilon_h^2)$ rounds, these rounds contribute the same regret order. Lemma~\ref{lem:b-event} bounds the regret from the events where these guarantees fail. Finally, using $\varepsilon_h=2^{-h/2}$ and summing over phases yields $\operatorname{Reg}_T=\widetilde O(\sqrt{r(S+K)T})$. The full proof is in Appendix~\ref{app:blocks}.

\FloatBarrier

\section{General Context Distributions}\label{sec:general-contexts}
In this section we extend the result for a uniform context distribution to arbitrary context distributions in two steps.

\subsection{Approximately Uniform Context Distributions}\label{sec:almost-uniform}
\begin{corollary}\label{cor:almost-uniform}
Suppose $1/(8S)\le\nu(i)\le8/S$ for all $i\in[S]$ and the induced block distribution is arbitrary. Algorithm~\ref{alg:arbitrary-blocks} satisfies \mbox{$\operatorname{Reg}_T=\widetilde O\!\left(\sqrt{r(S+K)T}\right)$}. The same holds when $r\le\min\{S,K\}$ is only an upper bound on the number of nonempty blocks; neither $\nu$ nor $T$ need be known.
\end{corollary}
Corollary~\ref{cor:almost-uniform} extends Theorem~\ref{thm:blocks} to approximately uniform context distributions. The lower bound on $\nu(i)$ ensures that Baseline and Collect receive sufficiently many observations, while the upper bound allows the average policy regret to remain $O(\varepsilon_h)$. Thus, the same regret rate is preserved up to constant factors. This extension allows Algorithm~\ref{alg:arbitrary-blocks} to be applied to groups of contexts with similar probabilities, which will be used to handle general context distributions.

The proof is in Appendix~\ref{app:almost-uniform}.

\subsection{Arbitrary Context Distributions}\label{sec:arbitrary-contexts}
For an arbitrary context distribution $\nu$, we adapt the frequency-based grouping approach of \citet[Algorithm~3 and Appendix~F]{lee2023context}. Algorithm~\ref{alg:general} partitions contexts according to their estimated arrival probabilities and runs Algorithm~\ref{alg:arbitrary-blocks} separately within each retained group. With probability at least $1-(T+1)^{-2}$, each group's conditional context distribution satisfies the assumptions of Corollary~\ref{cor:almost-uniform}, while the remaining low-frequency contexts have total probability at most $2\sqrt{S/T}$. For a given horizon $T$, define $J=\lceil32\sqrt{ST}\log(2S(T+1)^2)\rceil$. We may assume $J\le T$.

\begin{algorithm2e}[!htbp]
\caption{Online Algorithm for General Context Distributions}\label{alg:general}
Let $t$ denote the current time and initialize $J\leftarrow\lceil32\sqrt{ST}\log(2S(T+1)^2)\rceil$, $L\leftarrow1+\lceil\log_2\sqrt{ST}\rceil$\;
Play arm $1$ for $J$ rounds\;
$\widehat\nu(i)\leftarrow J^{-1}\sum_{t=1}^J\mathbf1\{i_t=i\}$ for $i\in[S]$\;
$\mathcal X_\ell\leftarrow\{i\in[S]:2^{-\ell-1}<\widehat\nu(i)\le2^{-\ell},\ \widehat\nu(i)\ge1/\sqrt{ST}\}$ for $\ell=0,\ldots,L-1$\;
$\mathcal X_L\leftarrow\{i\in[S]:\widehat\nu(i)<1/\sqrt{ST}\}$\;
\For{each nonempty $\mathcal X_\ell$ with $\ell<L$}{
Initialize Algorithm~\ref{alg:arbitrary-blocks} on $\mathcal X_\ell$, using\;
$S=|\mathcal X_\ell|$ and $r=\min\{r,|\mathcal X_\ell|\}$\;
}
\For{$t=J+1,\ldots,T$}{
Receive $i_t$\;
\eIf{$i_t\in\mathcal X_L$}{
Play arm $1$\;
}{
Find $\ell<L$ such that $i_t\in\mathcal X_\ell$\;
Use Algorithm~\ref{alg:arbitrary-blocks} for $\mathcal X_\ell$ to select an arm\;
Play the selected arm and observe the reward\;
Update only Algorithm~\ref{alg:arbitrary-blocks} for $\mathcal X_\ell$\;
}
}
\end{algorithm2e}
\FloatBarrier
\begin{theorem}\label{thm:general}
For an arbitrary context distribution $\nu$ and its induced block distribution $\omega$, Algorithm~\ref{alg:general} satisfies \mbox{$\operatorname{Reg}_T=\widetilde O\!\left(\sqrt{r(S+K)T}\right)$}. Restarting Algorithm~\ref{alg:general} on intervals of lengths $1,2,4,\ldots$ gives the same bound without knowing $T$.
\end{theorem}
The initial estimation rounds and the discarded contexts each contribute at most $\widetilde O(\sqrt{ST})$ expected regret. Applying Corollary~\ref{cor:almost-uniform} within each remaining group and summing by Cauchy--Schwarz gives the stated rate; see Appendix~\ref{app:general}.

\FloatBarrier

\section{Conclusions and Future Directions}
\label{sec:conclusion}

We establish expected regret $\widetilde O(\sqrt{r(S+K)T})$ for context-lumpable stochastic bandits under arbitrary block and context distributions, matching the lower bound up to logarithmic factors. Rather than recovering the unknown block partition, our approach constructs a shared candidate set $W_h$ containing a near-optimal arm for each block, and then selects a near-optimal arm for each context from this set. Together with early stopping to control exploration regret, this improves the dependence on $r$ in the previous online regret bound without requiring separation between blocks.

Context-lumpable bandits can be viewed as a special case of lumpable Markov decision processes (MDPs) with i.i.d.\ state transitions. Extending our approach to general lumpable MDPs, where transitions may depend on the state and action, is a natural direction for future work.

\section*{AI Disclosure}
OpenAI's ChatGPT and Codex were used to assist with language polishing, LaTeX formatting, compilation error checking. The authors take full responsibility for all theoretical results, proofs, and the final manuscript.

\bibliographystyle{plainnat}
\bibliography{reference}

\appendix
\section{Uniform Context Distribution and Arbitrary Block Distributions}\label{app:blocks}
All phase-$h$ probability bounds below also hold conditional on the observations and learner's random choices before round $t_h$. The letter $C$ denotes an absolute constant that may change between inequalities. For the analysis, recall
\begin{equation}\label{eq:block-accuracy}
\varepsilon_{h,b}=\max\left\{1,\frac{S}{r|\mathcal B(b)|}\right\}\varepsilon_h
=\max\left\{1,\frac1{r\omega(b)}\right\}\varepsilon_h.
\end{equation}
The second equality uses $\omega(b)=|\mathcal B(b)|/S$. This choice follows \citet[Appendix~G]{lee2023context}; Algorithm~\ref{alg:arbitrary-blocks} does not use $\varepsilon_{h,b}$ or the block distribution.

\subsection{Lemmas}\label{app:blocks-lemmas}
Algorithm~\ref{alg:arbitrary-blocks} ensures
\begin{equation}\label{eq:b-membership}
\pi_{h-1}(i)\in W_{h-1}\subseteq W_h\subseteq[K],\qquad
|W_h|\le\min\left\{K,1+\frac{rh(h+3)}2\right\}\le r(h+1)^2\le r\beta_h^2.
\end{equation}
Each phase $\ell$ adds at most $r(\ell+1)$ arms, so $|W_h|\le1+r\sum_{\ell=1}^h(\ell+1)$. The parameters give
\begin{gather*}
\beta_h\ge h+1,\qquad m_{h,n}\le9\beta_h/\varepsilon_n^2,\qquad
Sq_{h,n}\le\beta_hrK\varepsilon_n/\varepsilon_h+S,\\
\sum_{n=0}^h\varepsilon_n^{-2}=2^{h+1}-1\le2\varepsilon_h^{-2},\qquad
\sum_{n=0}^h\varepsilon_n^{-1}=\frac{2^{(h+1)/2}-1}{\sqrt2-1}\le4\varepsilon_h^{-1}.
\end{gather*}
In particular,
\begin{equation}\label{eq:b-sampling-budgets}
\sum_{n=0}^hSq_{h,n}m_{h,n}
\le\frac{36\beta_h^2rK+18S\beta_h}{\varepsilon_h^2},\qquad
\sum_{n=0}^hm_{h,n}\le\frac{18\beta_h}{\varepsilon_h^2}.
\end{equation}

\begin{lemma}\label{lem:b-baseline}
The output of Baseline with input $(h,\pi)$ satisfies $|B_h(i)-A(i,\pi(i))|\le\varepsilon_h/2$ for every $i\in[S]$, with probability at least
\[
1-2S(16SKr2^h)^{-20}-S(16SKr2^h)^{-160}.
\]
\end{lemma}

\begin{lemma}\label{lem:b-means}
With probability at least $1-2^{18}(16SKr2^h)^{-18}$, every empirical mean computed by Collect during phase $h$ satisfies
\[
|\widehat A_k(i,j)-A(i,j)|\le\sqrt{2\beta_h/k}.
\]
For each entry $j$ in $Q(i)$, $k\ge1$ counts its recorded rewards; the bound applies whenever that count is reached.
\end{lemma}

\begin{lemma}\label{lem:b-completion}
With probability at least $1-S[(r+1)(h+1)+1](16SKr2^h)^{-160}$, Collect ends with $a_i>|Q(i)|$ for every $i\in[S]$ each time it is used in phase $h$.
\end{lemma}

\begin{lemma}\label{lem:b-draws}
With probability at least $1-r(h+1)(16SKr2^h)^{-20}$, the random selections in Step 2 satisfy
\[
\exists i\in\mathcal B(b):j_b^\star\in Q_{h,n}(i)
\]
simultaneously for every $b\in[r]$ and $n\in\{0,\ldots,h\}$ for which $\varepsilon_n\ge\varepsilon_{h,b}$.
\end{lemma}

Let $\Omega_h$ be the event that the following hold simultaneously in phase $h$:
\begin{enumerate}\renewcommand{\labelenumi}{(\roman{enumi})}
\item Baseline returns $|B_h(i)-A(i,\pi_{h-1}(i))|\le\varepsilon_h/2$ for every $i\in[S]$.
\item Every empirical mean computed by Collect satisfies the inequality in Lemma~\ref{lem:b-means}.
\item Collect ends with $a_i>|Q(i)|$ for every $i\in[S]$ each time it is used.
\item For every $b,n$ with $\varepsilon_n\ge\varepsilon_{h,b}$, some $i\in\mathcal B(b)$ satisfies $j_b^\star\in Q_{h,n}(i)$ in Step 2.
\end{enumerate}

\begin{lemma}\label{lem:b-event}
For every phase $h$, $\Pr(\Omega_h)\ge1-(16SKr2^h)^{-8}$.
\end{lemma}

\begin{lemma}\label{lem:b-candidates}
On $\Omega_h$, Step 3 empties each $D_{h,n}$ within $r$ iterations. The resulting candidate set satisfies
\[
\mu(b,j_b^\star)-\max_{j\in W_h}\mu(b,j)
\le\min\{1,6\sqrt2\,\varepsilon_{h,b}\},\qquad b\in[r].
\]
\end{lemma}

\begin{lemma}\label{lem:b-policy}
On $\Omega_h$, the policy returned in phase $h$ satisfies
\begin{gather*}
\max_j A(i,j)-A(i,\pi_h(i))\le\min\{1,10\varepsilon_{h,b}\},
\qquad i\in\mathcal B(b),\ b\in[r],\\
\frac1S\sum_{i=1}^S[\max_j A(i,j)-A(i,\pi_h(i))]\le20\varepsilon_h.
\end{gather*}
These inequalities hold simultaneously with probability at least $1-(16SKr2^h)^{-8}$, regardless of the preceding policy's accuracy.
\end{lemma}

\subsection{Proof of Theorem~\ref{thm:blocks}}\label{app:blocks-theorem}
Using the parameter bounds preceding Lemma~\ref{lem:b-baseline} and $L_h(k)\le272Sk$ for $k\ge\beta_h$, Steps 1--4 take at most
\begin{align}
&L_h(m_h)+\sum_{n=0}^hL_h(q_{h,n}m_{h,n})
+r\sum_{n=0}^hL_h(m_{h,n})+L_h(|W_h|m_h)\notag\\
&\quad\le272S\left[m_h(1+|W_h|)+\sum_{n=0}^h(q_{h,n}+r)m_{h,n}\right]\notag\\
&\quad\le\frac{2^{14}\beta_h^3r(S+K)}{\varepsilon_h^2}.\label{eq:b-duration-bound}
\end{align}
The final loop fixes the phase length at
\begin{equation}\label{eq:b-duration}
 t_{h+1}-t_h=\left\lceil\frac{2^{16}\beta_h^3r(S+K)}{\varepsilon_h^2}\right\rceil.
\end{equation}

Fix $h\ge2$. On $\Omega_{h-1}$, Lemma~\ref{lem:b-policy} gives
\begin{equation}\label{eq:b-old-policy-average}
\frac1S\sum_{i=1}^S[\max_j A(i,j)-A(i,\pi_{h-1}(i))]
\le20\varepsilon_{h-1}<30\varepsilon_h.
\end{equation}
In Collect at level $n\le h$, fix one entry $j$ of $Q(i)$ that receives $k\ge1$ samples. If $k\ge2$ and $j\ne\pi_{h-1}(i)$, the stopping inequality was false after sample $k-1$. On $\Omega_h$, Lemmas~\ref{lem:b-baseline}--\ref{lem:b-means} give
\begin{align*}
(k-1)[A(i,\pi_{h-1}(i))-A(i,j)]
&\le(k-1)(3\varepsilon_n+\varepsilon_h/2)
+2\sqrt{2\beta_h(k-1)}\\
&\le\tfrac72m_{h,n}\varepsilon_n+2\sqrt{2\beta_hm_{h,n}}.
\end{align*}
For $k=1$ or $j=\pi_{h-1}(i)$, the left-hand side is zero. Using \eqref{eq:span} for the last sample and $k\le m_{h,n}\le9\beta_h/\varepsilon_n^2$ gives
\begin{align*}
&k[\max_{j'}A(i,j')-A(i,j)]\\
&\quad\le m_{h,n}[\max_{j'}A(i,j')-A(i,\pi_{h-1}(i))]
+\tfrac72m_{h,n}\varepsilon_n+2\sqrt{2\beta_hm_{h,n}}+1\\
&\quad\le\frac{9\beta_h}{\varepsilon_n^2}
[\max_{j'}A(i,j')-A(i,\pi_{h-1}(i))]
+\frac{41\beta_h}{\varepsilon_n}.
\end{align*}
Summing over one assignment per context and using \eqref{eq:b-old-policy-average}, we obtain, on $\Omega_{h-1}\cap\Omega_h$,
\begin{align}
&\sum_{i=1}^S\left(
\frac{9\beta_h}{\varepsilon_n^2}[\max_j A(i,j)-A(i,\pi_{h-1}(i))]
+\frac{41\beta_h}{\varepsilon_n}\right)\notag\\
&\quad\le S\beta_h\left(\frac{270\varepsilon_h}{\varepsilon_n^2}
+\frac{41}{\varepsilon_n}\right)
\le\frac{311S\beta_h}{\varepsilon_n}.\label{eq:b-all-contexts}
\end{align}
On $\Omega_{h-1}\cap\Omega_h$, summing \eqref{eq:b-all-contexts} over the entries tested in Steps 2--4 gives
\begin{align*}
\text{Step 2:}\quad&311S\beta_h\sum_{n=0}^h\frac{q_{h,n}}{\varepsilon_n}
\le C\frac{\beta_h^2rK(h+1)+S\beta_h}{\varepsilon_h},\\
\text{Step 3:}\quad&311Sr\beta_h\sum_{n=0}^h\varepsilon_n^{-1}
\le\frac{CSr\beta_h}{\varepsilon_h},\\
\text{Step 4:}\quad&\frac{311S\beta_h|W_h|}{\varepsilon_h}
\le\frac{CSr\beta_h^3}{\varepsilon_h}.
\end{align*}
Here Step 2 tests $q_{h,n}$ entries per context, Step 3 tests at most $r$ per level, and Step 4 tests $|W_h|\le r\beta_h^2$. We used $h+1\le\beta_h$ and $\sum_{n=0}^h\varepsilon_n^{-1}\le4/\varepsilon_h$.

All other plays use $\pi_{h-1}$ or $\pi_h$. Since $\Omega_{h-1}$ and $\pi_{h-1}$ are fixed before $t_h$, independence of context arrivals and \eqref{eq:b-old-policy-average} give
\begin{align*}
&\mathbb E\!\left[\mathbf1_{\Omega_{h-1}}
\sum_{t=t_h}^{t_{h+1}-1}\bigl(\max_j A(i_t,j)-A(i_t,\pi_{h-1}(i_t))\bigr)\right]\\
&\quad=\mathbb E\!\left[\mathbf1_{\Omega_{h-1}}
\frac{t_{h+1}-t_h}{S}\sum_i\bigl(\max_j A(i,j)-A(i,\pi_{h-1}(i))\bigr)\right]
\le30\varepsilon_h(t_{h+1}-t_h).
\end{align*}
This bounds the plays of $\pi_{h-1}$ after dropping $\mathbf1_{\Omega_h}$ and including all rounds of the phase. In the final loop, $\Omega_h$ and $\pi_h$ are determined before the next context arrives. The same calculation with Lemma~\ref{lem:b-policy} bounds those plays by $20\varepsilon_h(t_{h+1}-t_h)$. Therefore
\begin{align*}
&\mathbb E\!\left[\mathbf1_{\Omega_{h-1}\cap\Omega_h}
\sum_{t=t_h}^{t_{h+1}-1}(\max_j A(i_t,j)-A(i_t,j_t))\right]\\
&\quad\le\frac{C\beta_h^3r(S+K)}{\varepsilon_h}
+50\varepsilon_h(t_{h+1}-t_h)
\le\frac{C\beta_h^3r(S+K)}{\varepsilon_h}.
\end{align*}
By Lemma~\ref{lem:b-event}, $\Pr((\Omega_{h-1}\cap\Omega_h)^c)\le257(16SKr2^h)^{-8}$. On this complement, regret is at most the phase length by \eqref{eq:span}. Thus
\begin{align*}
&\mathbb E\!\left[\sum_{t=t_h}^{t_{h+1}-1}(\max_j A(i_t,j)-A(i_t,j_t))\right]\\
&\quad\le C\beta_h^3r(S+K)2^{h/2}
+\frac{257(t_{h+1}-t_h)}{(16SKr2^h)^8}.
\end{align*}
Since $(\log x)^3\le2x$ for $x\ge1$, we have $\beta_h^3\le16000(16SKr2^h)$. Consequently,
\begin{equation}\label{eq:b-bad-sum}
\sum_{h=2}^{\infty}\frac{257(t_{h+1}-t_h)}{(16SKr2^h)^8}
\le C\frac{r(S+K)}{(16SKr)^7}\sum_{h=2}^{\infty}2^{-6h}
=O(1).
\end{equation}

Let $H$ satisfy $t_H\le T<t_{H+1}$. Completing the last phase only increases regret, so
\[
\operatorname{Reg}_T
\le\min\{T,t_2-t_1\}
+Cr(S+K)\sum_{h=2}^H\beta_h^3 2^{h/2}+O(1).
\]
The first term satisfies
\[
\min\{T,t_2-t_1\}\le\sqrt{(t_2-t_1)T}
\le C\beta_1^{3/2}\sqrt{r(S+K)T}.
\]
If the sum is nonempty, phase $H-1$ is complete and
\[
2^{16}\beta_{H-1}^3r(S+K)2^{H-1}\le t_H-t_{H-1}<T,
\qquad\beta_H\le2\beta_{H-1}.
\]
It follows that
\begin{align*}
r(S+K)\sum_{h=2}^H\beta_h^3 2^{h/2}
&\le4\beta_H^3r(S+K)2^{H/2}\\
&\le\frac{4\beta_H^3}{2^{15/2}\beta_{H-1}^{3/2}}
\sqrt{r(S+K)T}
\le C\beta_H^{3/2}\sqrt{r(S+K)T}.
\end{align*}
The sum is zero for $H=1$. By \eqref{eq:b-duration}, $2^H\le2(T+1)$, so $\beta_H\le40\log(16SKr(T+1))$. Thus
\[
\operatorname{Reg}_T
\le C\log^{3/2}(16SKr(T+1))\sqrt{r(S+K)T}
=\widetilde O\!\left(\sqrt{r(S+K)T}\right).
\tag*{$\square$}
\]

\subsection{Proofs of the Lemmas}\label{app:blocks-lemma-proofs}
\paragraph{Proof of Lemma~\ref{lem:b-baseline}.}
Suppose Baseline or Collect begins at round $\tau$; condition on the observations and learner's random choices before $i_\tau$ is drawn. Fix $i$ and $j=\pi(i)$ for Baseline, or one entry $j$ of $Q(i)$ for Collect. For $k\ge1$, let $I_t$ indicate whether $y_t$ is among the first $k$ rewards recorded for this choice; sums below run over that algorithm's rounds.

Whether $I_t=1$ is determined before observing $y_t$. By \eqref{eq:noise} and iterated expectations,
\[
\mathbb E\exp\!\left(\lambda\sum_t I_t\eta_t-\frac{\lambda^2}{2}\sum_t I_t\right)\le1.
\]
With $\lambda=\sqrt{2\beta_h/k}$, Markov's inequality gives
\[
\Pr\!\left(\sum_t I_t=k,\ \sum_t I_t\eta_t>\sqrt{2\beta_hk}\right)
\le e^{-\beta_h}.
\]
When $\sum_t I_t=k$, $\sum_t I_t\eta_t=k(\widehat A_k(i,j)-A(i,j))$. Applying the same bound with $-\lambda$ yields
\begin{equation}\label{eq:adaptive-concentration}
\Pr\!\left(\sum_t I_t=k,\quad
|\widehat A_k(i,j)-A(i,j)|>\sqrt{2\beta_h/k}\right)\le2e^{-\beta_h}.
\end{equation}
Over $L_h(k)$ rounds starting at $\tau$, independence and the uniform context distribution give
\begin{align}
\Pr\!\left(\sum_{t=\tau}^{\tau+L_h(k)-1}\mathbf1\{i_t=i\}<k\right)
&\le e^k\left(1-\frac{1-e^{-1}}S\right)^{L_h(k)}\notag\\
&\le e^{k-L_h(k)/(2S)}\le e^{-8\beta_h}.\label{eq:uniform-arrivals}
\end{align}
Here $1-e^{-1}>1/2$ and $L_h(k)\ge128S(k+\beta_h)$. For Baseline, use $k=m_h$ and $\sqrt{2\beta_h/m_h}\le\varepsilon_h/2$. Equations~\eqref{eq:adaptive-concentration}--\eqref{eq:uniform-arrivals} and a union bound give
\[
\Pr\!\left(\exists i:\ |B_h(i)-A(i,\pi(i))|>\varepsilon_h/2\right)
\le2Se^{-\beta_h}+Se^{-8\beta_h}.
\tag*{$\square$}
\]

\paragraph{Proof of Lemma~\ref{lem:b-means}.}
Conditional on the choices of $Q(i)$, \eqref{eq:adaptive-concentration} applies to every entry and each $k\le m_{h,n}$. Counting repeated arms separately, Steps 2--4 compute at most
\begin{align*}
&S\sum_{n=0}^h(q_{h,n}+r)m_{h,n}+S|W_h|m_h\\
&\quad\le\frac{36\beta_h^2rK+18S\beta_h+18Sr\beta_h+9S|W_h|\beta_h}{\varepsilon_h^2}
\le\frac{64\beta_h^3r(S+K)}{\varepsilon_h^2}.
\end{align*}
The inequalities use \eqref{eq:b-sampling-budgets} and \eqref{eq:b-membership}. A union bound, $r(S+K)\le2SKr$, and $\beta_h^3\le16000(16SKr2^h)$ bound the failure probability by
\[
128\beta_h^3r(S+K)2^he^{-\beta_h}
\le\frac{256000}{(16SKr2^h)^{18}}
\le\frac{2^{18}}{(16SKr2^h)^{18}}.
\tag*{$\square$}
\]

\paragraph{Proof of Lemma~\ref{lem:b-completion}.}
In Collect at level $n$, each entry of $Q(i)$ uses at most $m_{h,n}$ samples. Over its $L_h(m_{h,n}\max_i|Q(i)|)$ rounds,
\[
\sum_t\mathbf1\{i_t=i\}\ge m_{h,n}|Q(i)|
\quad\Longrightarrow\quad a_i>|Q(i)|\ \text{at the end}.
\]
Equation~\eqref{eq:uniform-arrivals} bounds the failure probability by $Se^{-8\beta_h}$. Summing over the uses of Collect in Steps 2--4 gives
\[
S[(h+1)+r(h+1)+1]e^{-8\beta_h}
=S[(r+1)(h+1)+1](16SKr2^h)^{-160}.
\tag*{$\square$}
\]

\paragraph{Proof of Lemma~\ref{lem:b-draws}.}
Fix $b,n$ with $\varepsilon_n\ge\varepsilon_{h,b}$. The block has $|\mathcal B(b)|=S\omega(b)$ contexts. Thus
\begin{align*}
\Pr(j_b^\star\notin Q_{h,n}(i)\ \forall i\in\mathcal B(b))
&=(1-1/K)^{q_{h,n}|\mathcal B(b)|}\\
&\le\exp(-q_{h,n}|\mathcal B(b)|/K)
\le\exp(-\beta_hr\omega(b)\varepsilon_n/\varepsilon_h)
\le e^{-\beta_h},
\end{align*}
since $r\omega(b)\varepsilon_n/\varepsilon_h\ge r\omega(b)\varepsilon_{h,b}/\varepsilon_h=\max\{r\omega(b),1\}\ge1$. There are at most $r(h+1)$ relevant pairs $b,n$. A union bound gives the claim.\hfill$\square$

\paragraph{Proof of Lemma~\ref{lem:b-event}.}
Lemmas~\ref{lem:b-baseline}--\ref{lem:b-draws} give
\begin{align*}
\Pr(\Omega_h^c)
&\le\frac{2^{18}}{(16SKr2^h)^{18}}
+\frac{2S+r(h+1)}{(16SKr2^h)^{20}}
+\frac{S[(r+1)(h+1)+2]}{(16SKr2^h)^{160}}\\
&\le\frac{2^{18}}{(16SKr2^h)^{18}}
+\frac1{(16SKr2^h)^{19}}
+\frac1{(16SKr2^h)^{159}}
\le\frac1{(16SKr2^h)^8}.
\end{align*}
The second inequality follows from $h+1\le2^h$, and the last from $16SKr2^h\ge32$.\hfill$\square$

\paragraph{Proof of Lemma~\ref{lem:b-candidates}.}
On $\Omega_h$, if the stopping inequality holds at level $n$, then
\begin{align}
A(i,j)&\le\widehat A_k(i,j)+\sqrt{2\beta_h/k}<B_h(i)-3\varepsilon_n\notag\\
&\le A(i,\pi_{h-1}(i))+\varepsilon_h/2-3\varepsilon_n
\le A(i,\pi_{h-1}(i))-5\varepsilon_n/2.\label{eq:b-stop-implication}
\end{align}
An optimal arm cannot satisfy this inequality. By (iii)--(iv), whenever $\varepsilon_n\ge\varepsilon_{h,b}$, Step 2 records $(i,j_b^\star)\in D_{h,n}$ for some $i\in\mathcal B(b)$.

In Step 3, $(i,j^\star)\notin D$ implies $A(i,j^\star)<A(i,\pi_{h-1}(i))$ by (iii) and \eqref{eq:b-stop-implication}. Since completed estimates and $B_h$ have error at most $\varepsilon_n/2$, the definition of $\widehat V$ gives
\begin{equation}\label{eq:b-V}
|\widehat V(i,j^\star)-\max\{A(i,\pi_{h-1}(i)),A(i,j^\star)\}|\le\varepsilon_n/2.
\end{equation}
For the selected $(i^\star,j^\star)$ and any $i\in\mathcal B(g(i^\star))$,
\begin{align*}
\widehat V(i,j^\star)
&\ge A(i,j^\star)-\varepsilon_n/2
=A(i^\star,j^\star)-\varepsilon_n/2\\
&\ge\widehat A_{h,n}(i^\star,j^\star)-\varepsilon_n
>\widehat A_{h,n}(i^\star,j^\star)-4\varepsilon_n.
\end{align*}
All remaining records from this block are deleted. No records are added during Step 3, so each $D_{h,n}$ becomes empty within $r$ iterations.

For $\varepsilon_{h,b}\le1$, choose the largest $n\le h$ with $\varepsilon_n\ge\varepsilon_{h,b}$. Since $\varepsilon_0=1$ and $\varepsilon_{h,b}\ge\varepsilon_h$,
$\varepsilon_{h,b}\le\varepsilon_n<\sqrt2\,\varepsilon_{h,b}$.
Choose $(i,j_b^\star)\in D_{h,n}$ with $i\in\mathcal B(b)$. When this record is deleted,
\begin{align*}
\widehat A_{h,n}(i^\star,j^\star)
&\ge\widehat A_{h,n}(i,j_b^\star)
\ge\mu(b,j_b^\star)-\varepsilon_n,\\
\widehat V(i,j^\star)
&\ge\widehat A_{h,n}(i^\star,j^\star)-4\varepsilon_n.
\end{align*}
Since $j^\star,\pi_{h-1}(i)\in W_h$, \eqref{eq:b-V} implies
\begin{align*}
\max_{j\in W_h}\mu(b,j)
&\ge\max\{A(i,\pi_{h-1}(i)),A(i,j^\star)\}\\
&\ge\widehat V(i,j^\star)-\varepsilon_n
\ge\mu(b,j_b^\star)-6\varepsilon_n
\ge\mu(b,j_b^\star)-6\sqrt2\,\varepsilon_{h,b}.
\end{align*}
The gap is also at most one by \eqref{eq:span}. For $\varepsilon_{h,b}>1$, this alone suffices.\hfill$\square$

\paragraph{Proof of Lemma~\ref{lem:b-policy}.}
In Step 4, (iii) and \eqref{eq:b-stop-implication} ensure that every $j\in W_h$ with $A(i,j)\ge A(i,\pi_{h-1}(i))$ is recorded in $D$. Since $\pi_{h-1}(i)\in W_h$, this includes an arm maximizing $A(i,j)$ over $W_h$. Using the error bound $\varepsilon_h/2$ for all estimates, including $B_h$,
\begin{align*}
A(i,\pi_h(i))
&\ge\widehat A(i,\pi_h(i))-\varepsilon_h/2\\
&=\max_{j:(i,j)\in D}\widehat A(i,j)-\varepsilon_h/2
\ge\max_{j\in W_h}A(i,j)-\varepsilon_h.
\end{align*}
For $i\in\mathcal B(b)$, Lemma~\ref{lem:b-candidates} and \eqref{eq:span} give
\[
\max_j A(i,j)-A(i,\pi_h(i))
\le\min\{1,6\sqrt2\,\varepsilon_{h,b}+\varepsilon_h\}
\le\min\{1,10\varepsilon_{h,b}\}.
\]
Finally,
\begingroup
\postdisplaypenalty=10000
\begin{align*}
\frac1S\sum_i[\max_jA(i,j)-A(i,\pi_h(i))]
&\le10\sum_b\omega(b)\varepsilon_{h,b}
=10\varepsilon_h\sum_b\max\{\omega(b),1/r\}\\
&\le10\varepsilon_h\sum_b(\omega(b)+1/r)=20\varepsilon_h.
\end{align*}
Only $\Omega_h$ was used, so Lemma~\ref{lem:b-event} gives the stated probability.\hfill$\square$\par
\endgroup

\section{General Context Distributions}\label{app:general}
\allowdisplaybreaks[2]

\subsection{Proof of Corollary~\ref{cor:almost-uniform}}\label{app:almost-uniform}
\begin{proof}
Use the first definition in \eqref{eq:block-accuracy}, with sums over the nonempty blocks:
\[
\varepsilon_{h,b}=\max\left\{1,\frac{S}{r|\mathcal B(b)|}\right\}\varepsilon_h,
\qquad \sum_b\frac{|\mathcal B(b)|}{S}=1,\qquad\sum_b\frac1r\le1.
\]

Conditional on the history at a call's start $\tau$, $\nu(i)\ge1/(8S)$ gives
\begin{align*}
\Pr\!\left(\sum_{t=\tau}^{\tau+L_h(k)-1}\mathbf1\{i_t=i\}<k\right)
&\le e^k[1-(1-e^{-1})\nu(i)]^{L_h(k)}\\
&\le e^{k-L_h(k)/(16S)}
\le e^{-7k-8\beta_h}\le e^{-8\beta_h}.
\end{align*}
Together with \eqref{eq:adaptive-concentration}, this preserves Lemmas~\ref{lem:b-baseline}--\ref{lem:b-completion}. For $\varepsilon_n\ge\varepsilon_{h,b}$,
\begin{align*}
\Pr(j_b^\star\notin Q_{h,n}(i)\ \forall i\in\mathcal B(b))
&\le\exp\!\left(-\frac{q_{h,n}|\mathcal B(b)|}{K}\right)\\
&\le\exp\!\left(-\frac{\beta_hr|\mathcal B(b)|\varepsilon_n}{S\varepsilon_h}\right)
\le e^{-\beta_h}.
\end{align*}
Hence $\Pr(\Omega_h^c)\le(16SKr2^h)^{-8}$ as in Lemmas~\ref{lem:b-draws}--\ref{lem:b-event}. On $\Omega_h$, the same comparisons as in Lemmas~\ref{lem:b-candidates}--\ref{lem:b-policy} yield
\begin{align*}
\frac1S\sum_i[\max_jA(i,j)-A(i,\pi_h(i))]
&\le\frac{10}{S}\sum_b|\mathcal B(b)|\varepsilon_{h,b}\\
&=10\varepsilon_h\sum_b\max\{|\mathcal B(b)|/S,1/r\}
\le20\varepsilon_h,\\
\sum_i\nu(i)[\max_jA(i,j)-A(i,\pi_h(i))]
&\le\frac8S\sum_i[\max_jA(i,j)-A(i,\pi_h(i))]
\le160\varepsilon_h.
\end{align*}
For $h\ge2$, use \eqref{eq:b-all-contexts} for assigned plays. For plays of $\pi_{h-1}$, drop $\mathbf1_{\Omega_h}$ and condition at each call's start; for $\pi_h$, condition at the final loop's start. By \eqref{eq:b-membership} and \eqref{eq:b-duration},
\begin{align*}
&\mathbb E\!\left[\mathbf1_{\Omega_{h-1}\cap\Omega_h}
\sum_{t=t_h}^{t_{h+1}-1}(\max_j A(i_t,j)-A(i_t,j_t))\right]\\
&\quad\le CS\beta_h\left[\sum_{n=0}^h\frac{q_{h,n}+r}{\varepsilon_n}
+\frac{r\beta_h^2}{\varepsilon_h}\right]
+160(\varepsilon_{h-1}+\varepsilon_h)(t_{h+1}-t_h)\\
&\quad
\le\frac{C\beta_h^3r(S+K)}{\varepsilon_h}.
\end{align*}
For $t_H\le T<t_{H+1}$, \eqref{eq:b-bad-sum} and the phase summation in Appendix~\ref{app:blocks-theorem} now give
\begin{align*}
\operatorname{Reg}_T
&\le\min\{T,t_2-t_1\}
+Cr(S+K)\sum_{h=2}^H\beta_h^3 2^{h/2}+O(1)\\
&\le C\beta_H^{3/2}\sqrt{r(S+K)T}
\le C\log^{3/2}(16SKr(T+1))\sqrt{r(S+K)T}.
\end{align*}
\end{proof}

\subsection{Estimation of the Context Distribution}\label{app:partition}
\begin{lemma}\label{lem:context-partition}
Suppose $J<T$, with $J,L,\widehat\nu$ and the sets $\mathcal X_\ell$ defined in Section~\ref{sec:arbitrary-contexts}. With probability at least $1-(T+1)^{-2}$, both of the following hold:
\begin{enumerate}\renewcommand{\labelenumi}{(\roman{enumi})}
\item For every nonempty $\mathcal X_\ell$ with $\ell<L$ and every $i\in\mathcal X_\ell$,
\begin{equation*}
\frac1{8|\mathcal X_\ell|}
\le\frac{\nu(i)}{\sum_{i'\in\mathcal X_\ell}\nu(i')}
\le\frac8{|\mathcal X_\ell|}.
\end{equation*}
\item The remaining set has probability mass
\begin{equation*}
\sum_{i\in\mathcal X_L}\nu(i)\le2\sqrt{S/T}.
\end{equation*}
\end{enumerate}
\end{lemma}

\begin{proof}
Bernstein's inequality and a union bound give, simultaneously for all $i\in[S]$, with probability at least $1-(T+1)^{-2}$,
\begin{align*}
|\widehat\nu(i)-\nu(i)|
&\le\sqrt{\frac{2\nu(i)\log(2S(T+1)^2)}{J}}
+\frac{\log(2S(T+1)^2)}{J}\\
&\le\frac{\nu(i)}4+\frac{3\log(2S(T+1)^2)}{J}
\le\frac{\nu(i)}4+\frac3{32\sqrt{ST}}.
\end{align*}
Therefore
\[
\frac45\widehat\nu(i)-\frac3{40\sqrt{ST}}
\le\nu(i)\le\frac43\widehat\nu(i)+\frac1{8\sqrt{ST}}.
\]
For $i,i'\in\mathcal X_\ell$ with $\ell<L$,
\begin{align*}
\widehat\nu(i)\ge\frac1{\sqrt{ST}}
&\implies \frac12\widehat\nu(i)\le\nu(i)\le2\widehat\nu(i),\\
\frac{\nu(i)}{\nu(i')}
&\le4\frac{\widehat\nu(i)}{\widehat\nu(i')}\le8,\\
\frac1{8|\mathcal X_\ell|}
&\le\frac{\nu(i)}{\sum_{i'\in\mathcal X_\ell}\nu(i')}
\le\frac8{|\mathcal X_\ell|}.
\end{align*}
For the remaining contexts,
\[
\sum_{i\in\mathcal X_L}\nu(i)
\le\sum_{i\in\mathcal X_L}\left(\frac43\widehat\nu(i)+\frac1{8\sqrt{ST}}\right)
\le\frac{2|\mathcal X_L|}{\sqrt{ST}}
\le2\sqrt{S/T}.
\]
\end{proof}

\subsection{Proof of Theorem~\ref{thm:general}}\label{app:general-theorem}
\begin{proof}
If $J\ge T$, \eqref{eq:span} gives
\[
\operatorname{Reg}_T\le T\le J
\le33\sqrt{ST}\log(2S(T+1)^2)
\le C\log(16SKr(T+1))\sqrt{r(S+K)T}.
\]
For $J<T$, let $\mathcal E$ be the event in Lemma~\ref{lem:context-partition} and $T_\ell=\sum_{t=J+1}^T\mathbf1\{i_t\in\mathcal X_\ell\}$ for $0\le\ell\le L$. The groups and $\mathcal E$ depend only on the initial observations, so
\begin{align*}
\mathbb E[\mathbf1_{\mathcal E}T_L]
&\le(T-J)\mathbb E\!\left[\mathbf1_{\mathcal E}\sum_{i\in\mathcal X_L}\nu(i)\right]
\le2\sqrt{ST},\\
(T-J)\Pr(\mathcal E^c)&\le\frac{T-J}{(T+1)^2}\le1.
\end{align*}

On $\mathcal E$, condition on the initial history and subsequent group memberships until we average below; each $T_\ell$ is then fixed. For $i\in\mathcal X_\ell$ at a round assigned to copy $\ell<L$, independence of context arrivals gives
\[
\Pr(i_t=i\mid i_1,j_1,y_1,\ldots,i_{t-1},j_{t-1},y_{t-1})
=\frac{\nu(i)}{\sum_{i'\in\mathcal X_\ell}\nu(i')}
\in\left[\frac1{8|\mathcal X_\ell|},\frac8{|\mathcal X_\ell|}\right].
\]
Future memberships are independent of preceding rewards, and the current membership is fixed by $i_t$. Thus \eqref{eq:noise} remains valid, also for each copy's local history by iterated conditioning. Each nonempty group has at most $\min\{r,|\mathcal X_\ell|\}$ blocks, so Corollary~\ref{cor:almost-uniform} gives (trivially if $T_\ell=0$)
\begin{align*}
&\mathbb E\!\left[\sum_{\substack{J<t\le T\\i_t\in\mathcal X_\ell}}
\bigl(\max_j A(i_t,j)-A(i_t,j_t)\bigr)\right]\\
&\quad\le C\log^{3/2}(16SKr(T+1))\sqrt{r(|\mathcal X_\ell|+K)T_\ell}.
\end{align*}
Cauchy--Schwarz yields
\begin{align*}
\sum_{\ell<L}\sqrt{r(|\mathcal X_\ell|+K)T_\ell}
&\le\sqrt{r\left(\sum_{\ell<L}|\mathcal X_\ell|+LK\right)
\sum_{\ell<L}T_\ell}\\
&\le\sqrt{r(S+LK)(T-J)}.
\end{align*}
Averaging over the conditioning and using \eqref{eq:span},
\begin{align*}
\operatorname{Reg}_T
&\le J+\mathbb E\!\left[\mathbf1_{\mathcal E}\sum_{\ell<L}
\sum_{\substack{J<t\le T\\i_t\in\mathcal X_\ell}}
\bigl(\max_j A(i_t,j)-A(i_t,j_t)\bigr)\right]\\
&\qquad+\mathbb E[\mathbf1_{\mathcal E}T_L]+(T-J)\Pr(\mathcal E^c)\\
&\le J+2\sqrt{ST}+1
+C\log^{3/2}(16SKr(T+1))\sqrt{r(S+LK)T}\\
&\le C\log^2(16SKr(T+1))\sqrt{r(S+K)T}.
\end{align*}
The last step uses $J\le C\sqrt{ST}\log(16SKr(T+1))$ and $L\le C\log(16SKr(T+1))$.

Restarting on rounds $2^p,\ldots,2^{p+1}-1$, the same bound holds conditional on each restart's history. Complete the last interval and sum:
\begin{align*}
\operatorname{Reg}_T
&\le\sum_{p=0}^{\lfloor\log_2T\rfloor}\mathbb E\!\left[
\sum_{t=2^p}^{2^{p+1}-1}\bigl(\max_j A(i_t,j)-A(i_t,j_t)\bigr)\right]\\
&\le C\log^2(16SKr(T+1))\sqrt{r(S+K)}
\frac{2^{(\lfloor\log_2T\rfloor+1)/2}-1}{\sqrt2-1}\\
&\le C\log^2(16SKr(T+1))\sqrt{r(S+K)T}.
\end{align*}
\end{proof}

\end{document}